\documentclass[sigconf]{aamas}  

\usepackage{booktabs}

\usepackage{amsmath}
\usepackage{amssymb}
\usepackage{tikz}
\usetikzlibrary{backgrounds,patterns,calc}
\usepackage{pgfplots}
\usepackage{fancybox}
\usepackage{multirow}

\tikzset{
  treenode/.style = {shape=circle,
                     draw=black, thin, align=center,
                     top color=white, bottom color=blue!15!white},
  bnode/.style     = {treenode, font=\sc},
  onode/.style      = {treenode, font=\sc, bottom color=red!30},
  candi/.style={edge from parent/.style={green!50!black,very thick,draw,->,dashed,>=stealth,thick}},
  norm/.style={edge from parent/.style={solid,black,thin,draw,->,>=stealth}}
}

\usepackage{mathrsfs}

\newcommand{\styleset}[1]{\ensuremath{\mathcal{#1}}}

\newcommand{\statespace}[0]{\styleset{S}}
\newcommand{\transition}[0]{\styleset{T}}
\newcommand{\observation}[0]{\styleset{O}}
\newcommand{\observemodel}[0]{\styleset{Z}}
\newcommand{\operator}[0]{\styleset{A}}
\newcommand{\beliefspace}[0]{\styleset{B}}
\newcommand{\reachable}[0]{\styleset{R}}
\newcommand{\objective}[0]{\styleset{G}}
\newcommand{\dest}[0]{\mathit{Dest}}
\newcommand{\safe}[0]{\mathit{Safe}}

\newcommand{\satBSTitle}{Goal-Constrained Belief Space}
\newcommand{\satBS}{goal-constrained belief space}
\newcommand{\ipsAlgo}{BPS}
\newcommand{\figref}{Figure}
\newcommand{\algoref}{Algorithm}
\newcommand{\forref}{Formula}

\newcommand{\secref}{Section}
\newcommand{\defref}{Definition}

\usepackage[algoruled,vlined,linesnumbered,algo2e,noresetcount]{algorithm2e}

\SetCommentSty{mycommfont}

\setcopyright{ifaamas}  
\acmDOI{}  
\acmISBN{}  
\acmConference[AAMAS'18]{Proc.\@ of the 17th International Conference on Autonomous Agents and Multiagent Systems (AAMAS 2018)}{July 10--15, 2018}{Stockholm, Sweden}{M.~Dastani, G.~Sukthankar, E.~Andr\'{e}, S.~Koenig (eds.)}  
\acmYear{2018}  
\copyrightyear{2018}  
\acmPrice{}  

\begin{document}

\title{Bounded Policy Synthesis for POMDPs with Safe-Reachability Objectives}  

\subtitle{Robotics Track}


\author{Yue Wang}
\affiliation{%
  \institution{Department of Computer Science Rice University}
 \city{Houston} 
 \state{TX 77005}
 \country{USA}
}
\email{yw27@rice.edu}

\author{Swarat Chaudhuri}
\affiliation{%
  \institution{Department of Computer Science Rice University}
 \city{Houston} 
 \state{TX 77005}
 \country{USA}
}
\email{swarat@rice.edu}

\author{Lydia E. Kavraki}
\affiliation{%
  \institution{Department of Computer Science Rice University}
 \city{Houston} 
 \state{TX 77005} 
 \country{USA}
}
\email{kavraki@rice.edu}

\begin{abstract}  
  Planning robust executions under uncertainty is a fundamental
  challenge for building autonomous robots. Partially Observable
  Markov Decision Processes (POMDPs) provide a standard framework for
  modeling uncertainty in many applications. In this work, we study
  POMDPs with \emph{safe-reachability} objectives, which require that
  with a probability above some threshold, a goal state is eventually
  reached while keeping the probability of visiting unsafe states
  below some threshold. This POMDP formulation is different from the
  traditional POMDP models with optimality objectives and we show that
  in some cases, POMDPs with safe-reachability objectives can provide
  a better guarantee of both safety and reachability than the existing
  POMDP models through an example. A key algorithmic problem for
  POMDPs is \emph{policy synthesis}, which requires reasoning over a
  vast space of beliefs (probability distributions). To address this
  challenge, we introduce the notion of a \emph{\satBS{}}, which only
  contains beliefs reachable from the initial belief under desired
  executions that can achieve the given safe-reachability
  objective. Our method compactly represents this space over a
  \emph{bounded} horizon using symbolic constraints, and employs an
  \emph{incremental} Satisfiability Modulo Theories (SMT) solver to
  efficiently search for a valid policy over it. We evaluate our
  method using a case study involving a partially observable robotic
  domain with uncertain obstacles. The results show that our method
  can synthesize policies over large belief spaces with a small number
  of SMT solver calls by focusing on the \satBS{}.
\end{abstract}

%

\keywords{Planning under Uncertainty; Policies; Robotics; Formal
  Methods} 

\maketitle

\section{Introduction}
\label{intro}

Partially Observable Markov Decision Processes (POMDPs)
\cite{smallwood1973optimal} provide a principled mathematical
framework for modeling a variety of problems in the face of
uncertainty
\cite{Mohri:1997:FTL:972695.972698,Kaelbling:1998:PAP:1643275.1643301,durbin1998biological,Chatterjee:2016:SSA:3016100.3016355}. As
an example, in robotics, accounting for uncertainty is a fundamental
challenge for deploying autonomous robots in the physical world. Many
applications in uncertain robotic domains can be modeled as POMDP
problems
\cite{Kaelbling:1998:PAP:1643275.1643301,grady2015extending,7139019,Chatterjee:2016:SSA:3016100.3016355}.

\begin{figure}[t!]
  \centering
  \includegraphics[width=0.95\columnwidth]{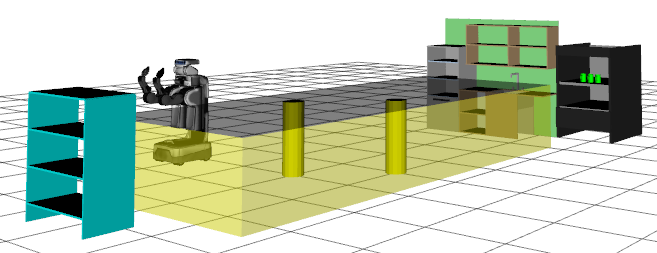}
  \caption{An example of a safe-reachability objective: a robot with
    uncertain actuation and perception needs to navigate through the
    kitchen and pick up a green cup from the black storage area
    (reachability), while avoiding collisions with uncertain obstacles
    (e.g., chairs) modeled as cylinders in the yellow ``shadow''
    region (safety). }
  \label{fig:kitchen}
\end{figure}

A key algorithmic problem for POMDPs is the synthesis of {\em
  policies} \cite{smallwood1973optimal}: recipes that specify the
actions to take under {\em all possible} events in the environment.
Typically, the goal in policy synthesis is to find optimal solutions
with respect to quantitative objectives such as maximizing discounted
reward
\cite{1307420,kurniawati2008sarsop,Bai2011,somani2013despot,Eck:2014:OHP:2615731.2615851,Grzes:2014:PPE:2615731.2615853,Grzes:2015:IPI:2772879.2773311,grady2015extending,luoimportance}.
While the purely quantitative formulations of the problem are suitable
for many applications, there are, settings that demand synthesis with
respect to {\em boolean} requirements. For example, consider the
scenario shown in \figref{} \ref{fig:kitchen} where we want to
guarantee that a robot can accomplish a task {\em safely} in an
uncertain domain. This goal is naturally formulated as policy
synthesis from a high-level requirement written in a temporal
logic. Moreover, in some cases, formulating boolean requirements as
quantitative objectives by assigning negative rewards for states that
violate the boolean requirements and positive rewards for states that
satisfy the boolean requirements, leads to policies that are overly
conservative or overly risky \cite{5509743}, depending on the
particular reward function chosen. Therefore, new models and
algorithms are required for handling POMDPs with boolean requirements
explicitly. In \secref{} \ref{sec_constrainedPOMDPs}, we discuss an
example that shows in some scenarios, handling boolean requirements
explicitly in POMDPs provides a better guarantee of both safety and
reachability than the traditional quantitative POMDP formulations.

Policy synthesis in POMDPs with respect to boolean requirements has
been studied before. Specifically, inspired by applications in
robotics, the \emph{qualitative analysis} problem of {\em almost-sure
  satisfaction} of POMDPs with temporal logic specifications was first
introduced in \cite{7139019}. In their work, the goal is to find
policies that satisfy a temporal property with probability 1.

A more general \emph{quantitative analysis} problem of POMDPs with
temporal logic specifications is to synthesize policies that satisfy a
temporal property with a probability above some threshold. In this
work, we study this problem for the special case of
\emph{safe-reachability} properties, which require that with a
probability above some threshold, a goal state is eventually reached
while keeping the probability of visiting unsafe states below some
threshold. Many robot tasks such as the one in \figref{}
\ref{fig:kitchen} can be formulated using a safe-reachability
objective.

Previous results
~\cite{Paz:1971:IPA:1097027,Madani:2003:UPP:873797.873799,Chatterjee2016878}
have shown that the quantitative analysis problem of POMDPs with
reachability objectives is undecidable. To make the problem tractable,
we assume there exists a \emph{bounded horizon} $h$ such that $h$ is
sufficiently large to prove the existence of a valid policy, or the
user is not interested in plans beyond the bounded horizon $h$. This
assumption is particularly reasonable for robotic domains because
robots are often required to accomplish a task in bounded steps due to
some resource constraints such as energy/time constraints. \figref{}
\ref{fig:kitchen} shows an example of such a scenario: a robot with
uncertain actuation and perception needs to navigate through a kitchen
to pick up an object in bounded steps while avoiding collisions with
uncertain obstacles.

To the best of our knowledge, the quantitative analysis problem of
POMDPs with safe-reachability objectives has not been considered
before. In this work, we present a practical policy synthesis approach
for this problem. Like most other algorithms for policy synthesis, our
approach is based on reasoning about the space of \emph{beliefs}, or
probability distributions over possible states of the POMDP. Our
primary algorithmic challenge is that the belief space is a vast,
high-dimensional space of probability distributions.

Our approach to this challenge is based on the new notion of a
\emph{\satBS{}}. This notion takes inspiration from recent advances in
point-based algorithms
\cite{Pineau:2003:PVI:1630659.1630806,kurniawati2008sarsop,somani2013despot,luoimportance}
for POMDPs with discounted reward objectives. These POMDP algorithms
exploit the notion of the \emph{reachable belief space}
$\reachable(b_{init})$ from an initial belief $b_{init}$ and compute
an approximately optimal policy over $\reachable(b_{init})$ rather
than the entire belief space. Similarly, we compute a valid policy
over a \emph{\satBS{}}, which contains beliefs visited by desired
executions that can achieve the safe-reachability objective. The
\satBS{} is generally much smaller than the original belief space.

Our synthesis algorithm, \emph{bounded policy synthesis} (\ipsAlgo{}),
computes a valid policy by iteratively searching for a candidate plan
in the \satBS{} and constructing a policy from this candidate plan. We
compactly represent the \satBS{} over a \emph{bounded} horizon using
symbolic constraints. The applicability of constraint-based methods
has been already advocated in several robotics planning algorithms
\cite{nedunuri2014smt,wang2016task,dantam2016tmp,kaelbling2016implicit}. Many
of these algorithms take advantage of a modern, incremental SMT solver
\cite{de2008z3} for efficiency. Inspired by this, we apply the SMT
solver to efficiently explore the symbolic \satBS{} to generate
candidate plans. Note that a candidate plan is a single path that only
covers a particular observation at each step, while a valid policy is
contingent on all possible observations. Therefore, once a candidate
plan is found, \ipsAlgo{} tries to generate a valid policy from the
candidate plan by considering all possible events at each step. If
this policy generation fails, \ipsAlgo{} adds additional constraints
that \emph{block} invalid plans and force the SMT solver to generate
other better plans. The incremental capability of the SMT solver
allows \ipsAlgo{} to efficiently generate alternate candidate plans
when we update the constraints. If there is no new candidate plan for
the current horizon, \ipsAlgo{} increases the horizon and repeats the
above steps until it finds a valid policy or reaches a given horizon
bound.

In summary, the contributions of the paper are:
\begin{itemize}
\item We show that in some domains, our formulation of POMDPs with
  safe-reachability objectives offers a better guarantee of both
  safety and reachability than the existing POMDP models through an
  example (\secref{} \ref{sec_constrainedPOMDPs}).
\item We introduce the notion of a \satBS{} to address the scalability
  challenge of solving POMDPs with safe-reachability objectives. Based
  on this notion, we present a novel approach called \ipsAlgo{} for
  policy synthesis of POMDPs with safe-reachability objectives.
\item We evaluate the scalability of \ipsAlgo{} using a case study
  involving a partially observable robotic domain with uncertain
  obstacles (\figref{} \ref{fig:kitchen}). The experimental results
  demonstrate that \ipsAlgo{} can scale up to huge belief spaces by
  focusing on the \satBS{}.
\end{itemize}

\section{Related Work}
\label{related_work}
POMDPs \cite{smallwood1973optimal} provide a principled mathematical
framework for modeling a variety of robotics problems in the face of
uncertainty. Many POMDP algorithms
\cite{luoimportance,kurniawati2008sarsop,Bai2011,somani2013despot,grady2015extending,1307420}
for robot applications focus on discounted reward objectives. Recent
work
\cite{Svorenova:2015:TLM:2728606.2728617,7139019,Chatterjee:2016:SSA:3016100.3016355}
has investigated \emph{almost-sure satisfaction} of POMDPs with
temporal logic specifications, where the goal is to check whether a
temporal logic objective can be ensured with probability 1. Our
approach can be seen as synthesizing policies for large POMDP problems
with basic temporal logic objectives (safe-reachability), but not
limited to almost-sure satisfaction analysis. Though we may formulate
a safe-reachability objective as an optimization problem by assigning
negative rewards for unsafe states and positive rewards for goal
states, this formulation does not always yield good policies
\cite{5509743}.

Recently, there has been a large body of work that extends the
traditional POMDP model with notions of risk and cost, including
constrained POMDPs (C-POMDPs)
\cite{Isom:2008:PLD:1619995.1620043,5509743,Kim:2011:PVI:2283696.2283730,Poupart:2015:ALP:2888116.2888181},
risk-sensitive POMDPs (RS-POMDPs)
\cite{Marecki:2010:RPP:1838206.1838384,Hou:2016:SRP:3016100.3016342}
and chance-constrained POMDPs (CC-POMDPs)
\cite{Santana:2016:RAC:3016100.3016366}. There are two major
differences between their models and our formulation of POMDPs with
safe-reachability objectives. First, the objective of these models is
to maximize the cumulative expected reward while keeping the expected
cost/risk below some threshold, while in our case, the objective is to
satisfy a safe-reachability objective in all possible executions
including the worst case, providing a better safety guarantee than the
formulation of expected cost/risk threshold constraints. Second,
C/RS/CC-POMDPs typically need to assign a proper positive reward for
goal states to ensure reachability and do not have direct control over
the probability of reaching goal states (e.g., reach a goal state with
a probability greater than some threshold), while our
safe-reachability objective can directly encode this probability
threshold constraint as a boolean requirement, providing a better
reachability guarantee than the quantitative formulation of
C/RS/CC-POMDPs.  While C/RS/CC-POMDPs are suitable for many
applications, there are domains in robotics such as autonomous driving
and disaster rescue that demand synthesis of policies that can provide
such strong guarantee of reaching goal states safely.

Task and Motion Planning (TMP)
\cite{6906922,hadfield2015modular,kaelbling2016implicit,erdem2011combining,gharbi2015combining,he2015towards,bidot2017geometric,wang2016task,dantam2016tmp}
describes a class of challenging problems that combine low-level
motion planning and high-level task reasoning. Most of these TMP
approaches focus on deterministic domains, while several of them apply
to uncertain domains with uncertainty in perception
\cite{hadfield2015modular,kaelbling2016implicit}. The main difference
is that, the above works perform \emph{online} planning with a
determinized approximation of belief space dynamics
\cite{platt2010belief} assuming the most likely observation will be
obtained, while our approach synthesizes a valid policy \emph{offline}
contingent on all possible events.


Our method computes a valid policy by iteratively searching for a
candidate plan that is likely to succeed with determinized
observations in the \satBS{}, and then constructing a policy from this
candidate plan by considering other possible observations. This idea
has been shown to improve the scalability of algorithms for a variety
of uncertain domains
\cite{dean1995planning,AAAI148656,ICAPS1613188}. The scalability of
our approach also relies on exploiting the notion of a
\emph{\satBS{}}. This idea resembles efficient point-based POMDP
algorithms \cite{luoimportance,kurniawati2008sarsop} based on
(optimally) reachable belief space.

We apply techniques from Bounded Model Checking (BMC)
\cite{biere2003bounded} to compactly represent the \satBS{} over a
bounded horizon. BMC verifies whether a finite state system satisfies
a given temporal logic specification. Thanks to the tremendous
increase in the reasoning power of practical SMT (SAT) solvers, BMC
can scale up to large systems with hundreds of thousands of
states. Our approach efficiently explores the \satBS{} by leveraging a
modern, incremental SMT solver \cite{de2008z3}. It has been shown that
the incremental capability of the SMT solver leads to an efficient
planning algorithm for TMP \cite{dantam2016tmp}. Inspired by this
result, we now leverage incremental SMT solvers for belief space
policy synthesis.

\section{Problem Formulation}

In this work, we consider the problem of policy synthesis for POMDPs:
\begin{definition}[POMDP]$ $\\
  A \emph{Partially Observable Markov Decision Process} (POMDP) is a
  tuple
  $P = (\statespace, \operator, \transition, \observation,
  \observemodel)$:
  \begin{itemize}
  \item $\statespace$ is a \emph{finite} set of states.
  \item $\operator$ is a \emph{finite} set of actions. 
  \item $\transition$ is a probabilistic transition function
    $\transition(s, a, s') = p(s'|s, a)$, which defines the
    probability of moving to state $s' \in \statespace$ after taking
    an action $a \in \operator$ in state $s \in \statespace$.
  \item $\observation$ is a \emph{finite} set of observations.
  \item $\observemodel$ is the probabilistic observation function
    $\observemodel(s', a, o) = p(o|s', a)$, which defines the
    probability of observing $o \in \observation$ after taking an
    action $a \in \operator$ and reaching state $s' \in \statespace$.
  \end{itemize}
\end{definition}

Due to uncertainty in transition and observation, the actual state is
partially observable and typically we maintain a \emph{belief}, which
is a probability distribution over all possible states
$b: \statespace \rightarrow [0, 1]$ with
$\sum\limits_{s \in \statespace}b(s) = 1$. The set of beliefs
$\beliefspace = \{b: \statespace \rightarrow [0, 1] ~| \sum\limits_{s
  \in \statespace}b(s)=1\}$ is known as \emph{belief space}. Note that
a transition $\transition_B$ in belief space is a \emph{deterministic}
function $b' = \transition_B(b, a, o)$, i.e., given an action
$a \in \operator$ and an observation $o \in \observation$, the updates
to beliefs are deterministic based on the formula:
\begin{equation}
\label{eq:belief_trans}
  b'(s') = \alpha\observemodel(s', a, o)\sum\limits_{s\in \statespace}\transition(s, a, s')b(s)
\end{equation}

where $\alpha$ is a normalization constant.

\begin{definition}[Plan] $ $\\
  A \emph{plan} in belief space is a sequence
  $\sigma = (b_0, a_1, o_1, b_1, a_2, o_2, b_2, \dots)$ such that for
  all $i>0$, the belief updates satisfy the transition function
  $\transition_B$, i.e., $b_i = \transition_B(b_{i-1}, a_i, o_i)$,
  where $a_i \in \operator$ is an action and $o_i \in \observation$ is
  an observation.
\end{definition}

\begin{definition}[Policy] $ $\\
  A \emph{policy} $\pi: \beliefspace \rightarrow \operator$ is a
  function that maps a belief $b \in \beliefspace$ to an action
  $a \in \operator$. A policy $\pi$ defines a set of plans in belief
  space:
  $\Omega_{\pi} = \{\sigma = (b_0, a_1, o_1, \dots) ~|~ \forall i>0,
  a_i = \pi(b_{i-1}) ~and~ o_i \in \observation \}$. For each plan
  $\sigma \in \Omega_{\pi}$, the action $a_i$ at each step $i$ is
  chosen by the policy $\pi$.
\end{definition}

\subsection{Safe-Reachability Objective}
In this work, we consider POMDPs with \emph{safe-reachability}
objectives:
\begin{definition}[Safe-Reachability Objective] $ $\\
  A \emph{safe-reachability objective} is a tuple $\objective = (\dest,\safe)$:
  \begin{itemize}
  \item $\safe$ is a set of safe beliefs
  \item $\dest$ is a set of goal beliefs.  In general, goal beliefs
    are safe beliefs, i.e., $\dest \subseteq \safe$.
  \end{itemize}
\end{definition}

A safe-reachability objective $\objective$ compactly represents the
set $\Omega_{\objective}$ of satisfiable plans in belief space:
\begin{definition}[Satisfiable Plan] $ $\\
  A plan $\sigma = (b_0, a_1, o_1, \dots)$ \emph{satisfies} a
  safe-reachability objective $\objective = (\dest, \safe)$ if there
  exists a belief $b_k$ at step $k$ in the plan $\sigma$ that is a
  goal belief $b_k \in \dest$ and all the beliefs $b_i$ ($i < k$)
  visited before step $k$ are safe beliefs $b_i \in \safe$.
\end{definition}

Note that safe-reachability objectives are defined using sets of
beliefs (probability distributions). The quantitative analysis problem
of POMDPs with requirements of a goal state is eventually reached with
a probability above some threshold while keeping the probability of
visiting unsafe states below some threshold, can be easily formulated
as a safe-reachability objective $\objective = (\dest,\safe)$ defined
as follows:
\begin{align}
  \dest & = \{b \in \beliefspace ~|~\left(\sum\limits_{s \text{ is a goal state}} b(s)\right) > 1 - \delta_1\} \\
  \safe & = \{b \in \beliefspace ~|~\left(\sum\limits_{s \text{ violates safety}} b(s)\right) < \delta_2\}
\end{align}

Where $\delta_1$ and $\delta_2$ are a small values that represents
tolerance.

\subsection{\satBSTitle}
\label{sec:sat}

It is intractable to compute a full policy that satisfies a given
safe-reachability objective for POMDPs, even under the assumption of
bounded horizon, due to the curse of dimensionality
\cite{Papadimitriou:1987:CMD:35577.35581}: the belief space
$\beliefspace$ is a high-dimensional, continuous space that contains
an infinite number of beliefs.

However, the \emph{reachable belief space} \cite{kurniawati2008sarsop}
$\reachable(b_{init})$ that contains beliefs reachable from the given
initial belief $b_{init}$, is much smaller than $\beliefspace$ in
general. Moreover, the safe-reachability objective $\objective$ defines a
set $\Omega_{\objective}$ of plans that satisfy $\objective$. Combining
$\reachable(b_{init})$ and $\Omega_{\objective}$, we can construct a
\emph{\satBS{}} $\reachable^{*}(b_{init}, \objective)$ that contains
beliefs reachable from the initial belief $b_{init}$ under satisfiable
plans $\sigma \in \Omega_{\objective}$. The \emph{\satBS{}}
$\reachable^*(b_{init}, \objective)$ is usually much smaller than the
reachable belief space $\reachable(b_{init})$. Thus, computing
policies over the \satBS{} $\reachable^*(b_{init}, \objective)$ can lead
to a substantial gain in efficiency.

\begin{figure}[t!]
  \centering
    \begin{tikzpicture}
      \tikzstyle{state}=[minimum width=20pt,draw=black,shape=circle,node distance=9em];
      \tikzstyle{action}=[minimum width=0pt,node distance=4em];
      \tikzstyle{a1_edge}=[->,>=stealth,thick,dashed,color=green!50!black];
      \tikzstyle{a2_edge}=[->,>=stealth,thick,color=red!50!black];

                    \node[state,name=s1,xshift=-2em] {$s_{ready}$};
                    \node[state,name=s2,above of=s1] {$s_{unsafe}$};
                    \node[state,name=s3,below of=s1] {$s_{goal}$};
                    \node[action,name=a1,left of=s1] {$a_{L}$};
                    \node[action,name=a2,right of=s1] {$a_{R}$};

                    \draw (s1.west) edge[out=150,in=-150,a1_edge]
                    node[left,align=center]
                    {0.1\\$(0.3,0.7)$}
                    (s2.west);

                    \draw (s1.west) edge[a1_edge,out=-150,in=150]
                    node[left,align=center]
                    {0.9\\$(0.8,0.2)$}
                    (s3.west);

                    \draw (s1) edge[loop below,a2_edge]
                    node[align=center] {0.05\\$(0.8, 0.2)$}
                    (s1);
                    
                    \draw (s1.east) edge[out=30,in=-30,a2_edge]
                    node[right,align=center]
                    {0.1\\$(0.8,0.2)$} (s2.east);
                    
                    \draw (s1.east) edge[out=-30,in=30,a2_edge]
                    node[right,align=center]
                    {0.85\\$(0.8,0.2)$} (s3.east);
    \end{tikzpicture}
    \caption{An example to show the difference between our formulation
      of POMDPs with safe-reachability objectives and
      unconstrained/C/RS/CC-POMDPs. There are 3 states: start state
      $s_{\textrm{ready}}$, unsafe state $s_{\textrm{unsafe}}$ and
      goal state $s_{\textrm{goal}}$. Dashed green edges represent
      transitions of executing left-hand pick-up action $a_L$ in state
      $s_{\textrm{ready}}$ and solid red edges represent transitions
      of executing right-hand pick-up action $a_{R}$ in state
      $s_{\textrm{ready}}$. For each edge, the first line is the
      transition probability and the second line is the tuple of
      observation probabilities
      $(p_{o_{\textrm{pos}}},p_{o_{\textrm{neg}}})$.}
    \label{fig_pomdp}
\end{figure}
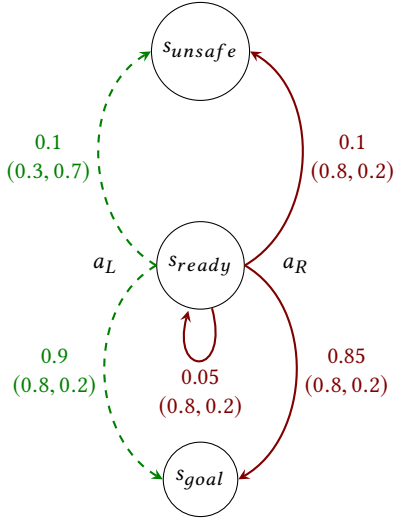

\subsection{Problem Statement}
Given a POMDP
$P = (\statespace, \operator, \transition, \observation,
\observemodel)$, an initial belief $b_{init}$ and a safe-reachability
objective $\objective$, our goal is to synthesize a \emph{valid} policy
$\pi_{\reachable^*}$ over the corresponding \satBS{}
$\reachable^*(b_{init}, \objective)$:
\begin{definition}[Valid Policy] $ $\\
  A \emph{valid} policy $\pi_{\reachable^*}: \reachable^*(b_{init},
  \objective) \mapsto
  \operator$ over a \satBS{} is a function that maps a belief $b \in
  \reachable^*(b_{init}, \objective)$ to an action $a \in
  \operator$. Therefore, the set
  $\Omega_{\pi_{\reachable^*}}$ of plans defined by the policy
  $\pi_{\reachable^*}$ is a subset of the set
  $\Omega_{\objective}$ defined by the safe-reachability objective
  $\objective$. i.e., $\Omega_{\pi_{\reachable^*}} \subseteq
  \Omega_{\objective}$.
  \label{def:policy}
\end{definition}

\section{Relation to Unconstrained POMDPs and C/RS/CC-POMDPs}
\label{sec_constrainedPOMDPs}

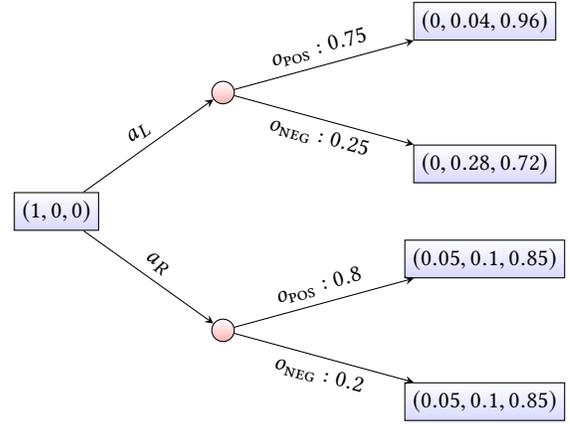
\begin{figure}[t!]
  \centering
  \begin{tikzpicture}
   [
    grow = right,
    sibling distance        = 6em,
    level distance          = 11em,
    edge from parent/.style = {draw=black,->,>=stealth,thick},
    every node/.style       = {font=\sc},
    sloped
    ]
    \node [bnode,xshift=-2em,,shape=rectangle] {$(1,0,0)$}
    child [norm] { node [onode,xshift=-4em,yshift=-2em] {}
      child [norm] { node [bnode,shape=rectangle] {$(0.05, 0.1, 0.85)$}
        edge from parent node [below] {$o_{\textrm{neg}}:0.2$}
      }
      child [norm] { node [bnode,shape=rectangle] {$(0.05, 0.1, 0.85)$}
        edge from parent node [above] {$o_{\textrm{pos}}:0.8$}
      }
      edge from parent node [above] {$a_R$}
    }
    child [norm] { node [onode,xshift=-4em,yshift=2em] {}
      child [norm] { node [bnode,shape=rectangle] {$(0, 0.28, 0.72)$}
        edge from parent node [below] {$o_{\textrm{neg}}:0.25$}
      }
      child [norm] { node [bnode,shape=rectangle] {$(0, 0.04, 0.96)$}
        edge from parent node [above] {$o_{\textrm{pos}}:0.75$}
      }
      edge from parent node [above] {$a_L$}
    };
    \end{tikzpicture}
    \caption{The belief space transition for the POMDP in \figref{}
      \ref{fig_pomdp}. Blue nodes
      $(p_{s_{\textrm{ready}}},p_{s_{\textrm{unsafe}}},p_{s_{\textrm{goal}}})$
      represent beliefs (probability distribution over states) and red
      nodes represent observation. The edges from blue nodes to red
      nodes represent actions and the edges from red nodes to blue
      nodes represent observations and the corresponding
      probabilities.}
    \label{fig_belief}
\end{figure}

There are two distinct approaches that can model safe-reachability
objectives \emph{implicitly} using the existing POMDP models in the
literature. The first approach is to incorporate safety and
reachability constraints as negative penalties for unsafe states and
positive rewards for goal states in \emph{unconstrained} POMDPs with
quantitative objectives. However, the authors of \cite{5509743} have
shown a counterexample that demonstrates formulating constraints as
unconstrained POMDPs with quantitative objectives does not always
yield good policies. The second approach is to encode
safe-reachability objectives implicitly as C/RS/CC-POMDPs that extend
unconstrained POMDPs with notions of risk and cost
\cite{Isom:2008:PLD:1619995.1620043,5509743,Marecki:2010:RPP:1838206.1838384,Kim:2011:PVI:2283696.2283730,Poupart:2015:ALP:2888116.2888181,Santana:2016:RAC:3016100.3016366,Hou:2016:SRP:3016100.3016342}. In
this section, we show the differences between POMDPs with
safe-reachability objectives and unconstrained/C/RS/CC-POMDPs through
an example.

In \figref{} \ref{fig:kitchen}, after the robot passes the yellow
``shadow'' region and moves to the position where it is ready to pick
up a green cup from the black storage area (start state
$s_{\textrm{ready}}$), it needs to decide how to pick up the
object. There are two action choices: pick-up using the left hand
(action $a_{L}$) and pick-up using the right hand (action
$a_{R}$). Both $a_L$ and $a_R$ are uncertain, and the robot may hit
the storage while executing $a_L$ or $a_R$, which results in an unsafe
collision state $s_{\textrm{unsafe}}$. There are two possible
observations after executing $a_L$ or $a_R$: observation
$o_{\textrm{pos}}$ representing the robot observes a cup in its hand
and observation $o_{\textrm{neg}}$ representing the robot observes no
cup in its hand (Note that the actual state may be different from the
observation due to uncertainty). The task objective is to reach a goal
state $s_{\textrm{goal}}$ where the robot holds a cup in its hand with
a probability greater than $0.8$ (reachability) while keeping the
probability of visiting unsafe state $s_{\textrm{unsafe}}$ below the
threshold $0.2$ (safety). The probability transition and observation
functions are shown in \figref{} \ref{fig_pomdp}. Based on \forref{}
\ref{eq:belief_trans}, we can get the transition in the corresponding
belief space (see \figref{} \ref{fig_belief}).

If we model this problem as an unconstrained POMDP by assigning a
negative penalty $-P$ ($P>0$) for unsafe state $s_{\textrm{unsafe}}$
and a positive reward $R$ ($R>0$) for goal state $s_{\textrm{goal}}$,
the optimal action for $s_{\textrm{ready}}$ that achieves the maximum
reward is always $a_L$, no matter what values of $P$ and $R$ are. This
is because the expected reward of action $a_L$ ($0.9R - 0.1P$) is
greater than the expected reward of $a_R$ ($0.85R - 0.1P$). However,
action $a_L$ does not satisfy the original safe-reachability objective
in the worst case where the robot observing $o_{\textrm{neg}}$ after
executing action $a_L$ and the resulting belief state
$(0, 0.28, 0.72)$ violates the original safety-reachability objective.

If we model this problem as a C/RS/CC-POMDP by assigning a positive
reward $R$ for goal state $s_{\textrm{goal}}$ and a cost $1$ for
visiting unsafe state $s_{\textrm{unsafe}}$, the best action for
$s_{\textrm{ready}}$ will be $a_{L}$ since both $a_L$ and $a_R$
satisfies the cost/risk constraint (expected cost/risk $0.1 < 0.2$)
and the expected reward of $a_L$ ($0.9R$) is greater than the expected
reward of $a_R$ ($0.85R$). However, action $a_L$ violates the original
safe-reachability objective for the same reason explained above.

On the other hand, using our formulation of POMDPs with
safe-reachability objectives, the best action for $s_{\textrm{ready}}$ will be
$a_R$. This is because, as shown in \defref{} \ref{def:policy}, a
valid policy in our formulation should satisfy the safe-reachability
objective in all possible executions and only $a_R$ satisfies the
safe-reachability objective in every possible execution.

The intent of this simple example is to illustrate that in some
domains where we want the robot to safely accomplish the task, our
formulation of POMDPs with safe-reachability objectives can provide a
better guarantee of both safety and reachability than the existing
POMDP models. While the formulations of cost/risk as negative
penalties in unconstrained POMDPs and expected cost/risk threshold
constraints in C/RS/CC-POMDPs are suitable for many applications,
there are domains such as autonomous driving and disaster rescue that
demand synthesis of policies that can provide such strong guarantee of
reaching goal states safely as in our formulation, especially when
violating safety requirements results in irreversible damage to
robots.

\section{Bounded Policy Synthesis}
\label{sec_synthesis}

\begin{figure}[t!]
      \centering
      \begin{tikzpicture}
        \tikzstyle{succ} = [color=green!50!black] 
        \tikzstyle{fail} = [color=red!50!black] 
        \tikzstyle{input} = [font=\small\sc\it,align=center];

        \tikzstyle{thing} = [font=\small\sc\sc,align=center, node
        distance=8em, draw=black,thick,rounded
        corners,fill=blue!15!white]; 
        \tikzstyle{edge} = [->,>=stealth,thick];

        \node[thing,name=cg]{Constraint\\Generation};
        \node[thing,name=candi,right of=cg,xshift=-1em]{Plan\\Generation};
        \node[thing,name=pr,right of=candi,xshift=-1em]{Policy\\Generation};

        \draw[edge] ($(cg.north) + (0,2em)$) node[above,input]
        {POMDP and\\Safe-Reachability\\Objective} --
        (cg.north);

        \draw (cg.north east)
        edge[out=30,in=150,edge,succ]
        node[above,input,succ] {constraints $\Phi_k$}
        (candi.north west);

        \draw (candi.south west)
        edge[out=210,in=-30,edge,fail]
        node[below,input,fail] {no new plan\\increase horizon}
        (cg.south east);

        \draw (candi.north east)
        edge[out=30,in=150,edge,succ]
        node[above,input,succ] {candidate\\plan}
        (pr.north west);

        \draw (pr.south west)
        edge[out=210,in=-30,edge,fail]
        node[below,input,fail] {additional\\constraints}
        (candi.south east);
       
        \draw(pr.east) edge[edge,succ] ++(1em,0)
        node[right,input,xshift=1em,succ] {Policy};

      \end{tikzpicture}
      \caption{The core steps of the \ipsAlgo{} algorithm.}
  \label{fig_algo}
\end{figure}
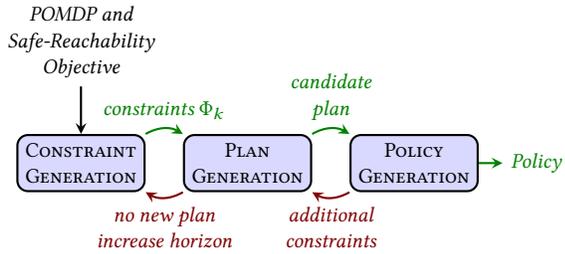

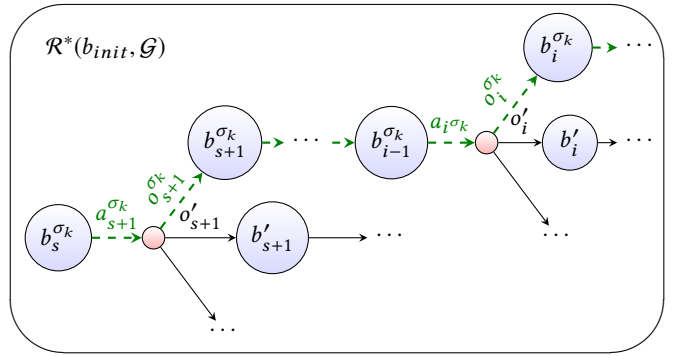
\begin{figure}[t!]
  \centering
  \ovalbox{
  \begin{minipage}{0.98\columnwidth}
    \centering
  \begin{tikzpicture}
  [
    grow = right,
    sibling distance        = 4em,
    level distance          = 3em,
    edge from parent/.style = {draw=black,->,>=stealth,thick},
    every node/.style       = {font=\sc},
    sloped
  ]
  \node [bnode,xshift=-2em] {$b_s^{\sigma_k}$}
    child [candi] { node [onode,xshift=1em] {}
      child [norm] { node  {$\cdots$}
      }
      child [norm] { node [bnode,solid,xshift=2em] {$b_{s+1}'$}
        child {node[xshift=2em] {$\cdots$}}
        edge from parent node [above] {$o_{s+1}'$}
      }
      child [candi] { node [bnode,solid] {$b_{s+1}^{\sigma_k}$}
        child {node[xshift=0.5em] {$\cdots$}
          child {node [bnode,solid,xshift=0.5em] {$b_{i-1}^{\sigma_k}$}
            child {node [onode,solid,xshift=1em] {}
              child [norm] { node  {$\cdots$}
              }
              child [norm] { node [bnode,solid,xshift=0.5em] {$b_{i}'$}
                child {node {$\cdots$}}
                edge from parent node [above] {$o_{i}'$}
              }
              child [candi] { node [bnode,solid] {$b_{i}^{\sigma_k}$}
                child {node[xshift=0.5em] {$\cdots$}}
                edge from parent node [above] {$o_{i}^{\sigma_k}$}
              }
              edge from parent node [above] {$a_{i^{\sigma_k}}$}
            }
          }
        }
        edge from parent node [above] {$o_{s+1}^{\sigma_k}$}
      }
      edge from parent node [above] {$a_{s+1}^{\sigma_k}$}
    }; 

    \node [font=\normalsize] at(0,8em) {$\reachable^*(b_{init}, \objective)$};
  \end{tikzpicture}
\end{minipage}
}
\caption{An example run of \ipsAlgo{}. The black box represents the
  \satBS{} $\reachable^*(b_{init}, \objective)$ over the bounded
  horizon $k$, blue nodes represent beliefs, red nodes represent
  observations, and the dashed green path represents one candidate
  plan $\sigma_k$ found by the incremental SMT solver. \ipsAlgo{}
  constructs a policy tree from this candidate plan by considering
  other branches following the red observation node for each step.}
  \label{fig:ips_ex}
\end{figure}

\begin{algorithm2e}[h!]
  \KwIn{\\{POMDP}
    $P = (\statespace, \operator, \transition, \observation,
    \observemodel)$\\
    {Initial Belief} $b_{init}$\\{Safe-Reachability Objective} $\objective$\\
    {Start Step} $s$\\{Horizon Bound} $h$} \KwOut{A
    \emph{Valid} Policy $\pi$}
  
  $k \leftarrow s$\tcc*{Initial horizon}

  $\Phi_k \leftarrow (b_s = b_{init})$ \label{line:init} \tcc*{Initial belief}
  
  \While(\label{line:bound}){$k \leq h$}{
    $\sigma_k \leftarrow \emptyset$\tcc*{$\sigma_k$: Candidate plan}
    \tcc{Add transition at step $k$ if $k > s$}
    \If{$k > s$} {
      $\Phi_k \leftarrow \Phi_k ~\wedge~ (b_{k} = \transition_B(b_{k-1}, a_{k},
      o_{k}))$\; \label{line:trans}
    }

    $\textrm{push}(\Phi_k)$\label{line:push}\tcc*{Push scope}
    \tcc{Add goal constraints at step $k$ (\forref{} \ref{liveness})}
    $\Phi_k \leftarrow \Phi_k ~\wedge~ G(\sigma_k, \objective,
      k)$; \label{line:goal} 

    \While{$\emptyset = \sigma_k$
      \label{line:gen_candi}
    }{
      \tcc{Candidate generation}
      $\sigma_k \leftarrow \textrm{IncrementalSMT}(\Phi_k)$\label{line:smt}\;

      \eIf(\tcc*[h]{No new plan}){$\emptyset = \sigma_k$} {
        \bf{break}\;
        }
        { 

          \tcc{$\phi$: constraints for blocking invalid plans}
          
          $\pi, \phi = \textrm{PolicyGeneration}(P,\objective,
          \sigma_k, s+1, k)$\; \label{line:pg}
          
            \eIf(\tcc*[h]{Generation failed}){$\emptyset \neq \phi$}{
              $\Phi_k \leftarrow \Phi_k ~\wedge~
              \phi$; \label{line:addCons} } { \Return
              $\pi$\;\label{line:succ} }
            $\sigma_k \leftarrow \emptyset$\; } } \tcc{Pop scope: pop
          goal and $\phi$ at step $k$}
        $\textrm{pop}(\Phi_k)$;\label{line:pop}

      $k \leftarrow k + 1$ \label{line:inc}
      \tcc*{Increase horizon} } \Return $\emptyset$\;
  \caption{\ipsAlgo{}}
  \label{alg:ips}
\end{algorithm2e}

The core steps of \ipsAlgo{} (\algoref ~\ref{alg:ips}) are shown in
\figref{} \ref{fig_algo}. \ipsAlgo{} computes a valid policy by
iteratively searching for a candidate plan in the \satBS{}
$\reachable^*(b_{init}, \objective)$ and constructing a valid policy
from this candidate plan. \figref{} \ref{fig:ips_ex} graphically
depicts one example run of \ipsAlgo{}.

First \ipsAlgo{} compactly encodes the \satBS{}
$\reachable^*(b_{\textrm{init}}, \objective)$ (the black box in \figref{}
\ref{fig:ips_ex}) w.r.t. the given POMDP
$P = (\statespace, \operator, \transition, \observation,
\observemodel)$, the initial belief $b_{\textrm{init}}$ and the
safe-reachability objective $\objective$ over a bounded \emph{horizon}
$k$ as a logical formula $\Phi_k$ (\algoref ~\ref{alg:ips}, lines
\ref{line:init}, \ref{line:trans}, \ref{line:goal}). We describe the
details of the constraints that encode the \satBS{} in \secref{}
\ref{sec:cg}.

Then \ipsAlgo{} computes a candidate plan by checking the
satisfiability of the constraint $\Phi_k$ (line \ref {line:smt})
through a modern, incremental SMT solver \cite{de2008z3}. Note that
the horizon $k$ restricts the length of the plan and thus the robot
can only execute $k$ actions.

If $\Phi_k$ is satisfiable, the SMT solver returns a candidate plan
(the dashed green path in \figref{} \ref{fig:ips_ex}) and \ipsAlgo{}
tries to generate a valid policy from the candidate plan by
considering all possible observations, i.e., other branches following
the red observation node at each step (line \ref{line:pg}). If this
policy generation succeeds, we find a valid policy. Otherwise,
\ipsAlgo{} adds additional constraints that block this invalid plan
(line \ref{line:addCons}) and forces the SMT solver to generate
another better candidate.

If $\Phi_k$ is unsatisfiable and thus there is no new plan for the
current horizon, \ipsAlgo{} increases the horizon by one (line
\ref{line:inc}) and repeats the above steps until a \emph{valid}
policy is found (line \ref{line:succ}) or a given horizon bound $h$ is
reached (line \ref{line:bound}).

This incremental SMT solver \cite{de2008z3} can efficiently generate
alternate candidate plans by maintaining a \emph{stack} of
\emph{scopes}, where each scope is a container for a set of
constraints and the corresponding ``knowledge'' learned from this set
of constraints. For fast repeated satisfiability checks, when we
update constraints (lines \ref{line:init}, \ref{line:trans},
\ref{line:goal}, \ref{line:addCons}), rather than rebuilding the
``knowledge'' from scratch, the incremental SMT solver only changes
the ``knowledge'' related to the updates by pushing (line
\ref{line:push}) and popping (line \ref{line:pop}) scopes. Thus the
``knowledge'' learned from previous satisfiability checks can be
reused.
 
\subsection{Constraint Generation}
\label{sec:cg}

In the first step, we use an encoding from Bounded Model Checking (BMC)
\cite{biere2003bounded} to construct the constraint $\Phi_k$
representing the \satBS{} $\reachable^*(b_{\textrm{init}}, \objective)$ w.r.t. the
POMDP
$P = (\statespace, \operator, \transition, \observation,
\observemodel)$, the initial belief $b_{\textrm{init}}$ and the
safe-reachability objective $\objective$ over the bounded horizon
$k$. The idea behind BMC is to find a finite plan with increasing
horizon that satisfies the given safe-reachability objective.

The constraint $\Phi_k$ contains three parts:
\begin{enumerate}
\item Starting from the initial belief (line \ref{line:init}):
  $b_s = b_{\textrm{init}}$.
\item Unfolding of the transition up to the horizon $k$ (line
  \ref{line:trans}):
  $\bigwedge_{i = s+1}^{k}(b_{i} = \transition_B(b_{i-1}, a_{i},
  o_{i}))$.
\item Satisfying the safe-reachability objective $\objective$ (line
  \ref{line:goal}).
\end{enumerate}

We can translate a safe-reachability objective to the constraint
$G(\sigma_k, \objective, k)$ on bounded plans
$\sigma_k = (b_s, a_{s+1}, o_{s+1}, \dots, a_k, o_k, b_k)$ using the
rules provided by BMC \cite{biere2003bounded} as follows:
\begin{equation}
  G(\sigma_k, \objective, k) = \bigvee_{i=s}^{k}(b_i \in \dest \wedge (\bigwedge_{j = s}^{i - 1} (b_j \in \safe))) 
\label{liveness}
\end{equation}

For a safe-reachability objective $\objective$ with a set $\dest$ of
goal beliefs and a set $\safe$ of safe beliefs, a finite plan that
visits a goal belief while staying in the safe region is sufficient to
satisfy $\objective$. Therefore, we only need to specify that a
bounded plan with length $k$ eventually visits a belief $b_i \in \dest$
while staying in the safe region
$(\bigwedge_{j = s}^{i - 1} (b_j \in \safe))$, as shown in \forref
~\ref{liveness}.

\subsection{Plan Generation}

The next step is to generate a candidate plan $\sigma_k$ of length $k$
that satisfies the constraint $\Phi_k$. We apply an incremental SMT
solver to efficiently search for such a candidate in the
\emph{\satBS{}} $\reachable^*(b_{\textrm{init}},\objective)$ defined by $\Phi_k$
(line \ref{line:smt}). If $\Phi_k$ is unsatisfiable, there is no
bounded plan $\sigma_k$ for the current horizon. In this case, we need
to increase the horizon (line \ref{line:inc}). If $\Phi_k$ is
satisfiable, the SMT solver will return a satisfying model that
assigns concrete values $b_i^{\sigma_k}$, $a_{i+1}^{\sigma_k}$ and
$o_{i+1}^{\sigma_k}$ for the belief $b_i$, action $a_{i+1}$ and
observation $o_{i+1}$ at each step $i$ respectively, which can be used
to construct the candidate plan
$\sigma_k = (b_s^{\sigma_k}, a_{s+1}^{\sigma_k}, o_{s+1}^{\sigma_k},
b_{s+1}^{\sigma_k} \dots, a_k^{\sigma_k}, o_k^{\sigma_k},
b_k^{\sigma_k})$.

\subsection{Policy Generation}

\begin{algorithm2e}[t!]
  \KwIn{\\
    {POMDP}
    $P = (\statespace, \operator, \transition, \observation,
    \observemodel)$\\ {Safe-Reachability Objective}
    $\objective$\\
    {Candidate Plan}
    $\sigma_k = (b_s^{\sigma_k}, a_{s+1}^{\sigma_k}, o_{s+1}^{\sigma_k},
    b_{s+1}^{\sigma_k} \dots)$\\
    {Start Step} $s$\\
    {Horizon Bound} $h$}

  \KwOut{A \emph{Valid} Policy $\pi$ {and Constraints $\phi$ for
      blocking invalid plans if the input candidate plan is invalid}}

  $\pi \leftarrow \emptyset$;
  
  \For(\label{line:scan}){$i = h$ \bf{downto} $s$}
  { 
    
    \ForEach(\label{line:enum}){observation
      $o \in \observation - \{o_i^{\sigma_k}\}$}{
      
      \tcc{Try observation $o$}
      $b_i' \leftarrow \transition_B(b_{i-1}^{\sigma_k},
      a_i^{\sigma_k}, o)$;\label{line:next}
      
      \tcc{Call \ipsAlgo{} to construct the branch}
      $\pi' \leftarrow \text{\ipsAlgo{}}(P, b_i', \objective, i,
      h)$; \label{line:branch}

       \If(\label{line:pgfail}\tcc*[h]{Construction failed})
       {$\emptyset = \pi'$} {
        
         Construct $\phi$ using \forref~\ref{eq:block}\label{line:block}

        \Return $\emptyset$, $\phi$\;
      }
      $\pi \leftarrow \pi \cup \pi'$\label{line:combine}\tcc*{Combine policy}
    }
    
    \tcc{Record action choice for belief $b_{i-1}^{\sigma_k}$}
    $\pi(b_{i - 1}^{\sigma_k}) \leftarrow a_i^{\sigma_k}$; \label{line:record}
  }
  \Return $\pi$, $\emptyset$\;

\caption{PolicyGeneration}
  \label{alg:pg}
\end{algorithm2e}

After \emph{plan generation}, we get a candidate plan $\sigma_k$ (the dashed
green path in \figref{} \ref{fig:ips_ex}) that satisfies the
safe-reachability objective $\objective$. This candidate plan is a single
path that only covers a particular observation $o_i^{\sigma_k}$ at
each step $i$. To construct a \emph{valid} policy, we should also
consider other possible observations $o_i' \neq o_i^{\sigma_k}$, i.e.,
other branches following the red observation node for each step
$i$. \emph{Policy generation} (\algoref~\ref{alg:pg}) tries to
construct a valid policy from a candidate plan by considering all
possible observations at each step.

For a candidate plan $\sigma_k$, we process each step of $\sigma_k$,
starting from the last step (\algoref ~\ref{alg:pg}, line
\ref{line:scan}). For each step $i$, since the set of observations
$\observation$ is finite, we can enumerate every possible observation
$o_i' \neq o_i^{\sigma_k}$ (line \ref{line:enum}) and compute the next
belief $b_i'$ using the transition function (line \ref{line:next}). To
ensure the action $a_i^{\sigma_k}$ also works for this different
observation $o_i'$, we need to compute a \emph{valid} policy for the
branch starting from $b_i'$, which is another \ipsAlgo{} problem and
can be solved using \algoref ~\ref{alg:ips} (line \ref{line:branch}).

If we successfully construct the valid policy $\pi'$ for this branch, we can
add $\pi'$ to the policy $\pi$ for the original synthesis problem
(line \ref{line:combine}). Otherwise, this candidate plan $\sigma_k$
can not be an element of a \emph{valid} policy
$\sigma_k \not\in \Omega_{\pi}$. In this case, we know that the prefix
of the candidate plan
$(b_s^{\sigma_k}, a_{s+1}^{\sigma_k}, o_{s+1}^{\sigma_k}, \dots,
b_{i-1}^{\sigma_k}, a_i^{\sigma_k})$ is invalid for current horizon
$k$ and we can add additional constraints $\phi$ to block all invalid
plans that have this prefix (line \ref{line:block}):
\begin{align}
  \phi = & ~\neg ~((b_s = b_s^{\sigma_k}) \wedge (a_i = a_i^{\sigma_k}) ~\wedge \nonumber\\
  & ~\left(\bigwedge_{m = s+1}^{i -1}
  (a_{m} = a_{m}^{\sigma_k}) \wedge (o_{m} = o_{m}^{\sigma_k})
  \wedge (b_m = b_m^{\sigma_k})\right)
  )
\label{eq:block}
\end{align}

Note that $\phi$ is only valid for current horizon $k$ and when we
increase the horizon, we should \emph{pop} the scope related to the
additional constraints $\phi$ from the stack of the SMT solver (line
\ref{line:pop}) so that we can \emph{revisit} this prefix with the
increased horizon. If we successfully construct policies for all other
branches at step $i$, we know that the choice of action
$a_i^{\sigma_k}$ for belief $b_{i-1}^{\sigma_k}$ is valid for all
possible observations. Then we record this choice for belief
$b_{i-1}^{\sigma_k}$ in the policy (line \ref{line:record}). This
policy generation terminates when it reaches the start step $s$ as
stated in the for-loop (line \ref{line:scan}) or it fails to construct
the valid policy $\pi'$ for a branch (line \ref{line:pgfail}).

\subsection{Algorithm Complexity}

\label{sec:complexity}

The \emph{reachable belief space} $\reachable(b_{\textrm{init}})$ can
be seen as a tree where the root node is the initial belief
$b_{\textrm{init}}$ and at each node, the tree branches on every
action and observation. The given horizon bound $h$ limits the height
of the tree. Therefore, the reachable belief space
$\reachable_h(b_{\textrm{init}})$ of height $h$ contains
$O(|\operator|^h|\observation|^h)$ plans, where $|\operator|$ and
$|\observation|$ are the size of action set $\operator$ and the size
of observation set $\observation$ respectively. To synthesize a
\emph{valid} policy, a naive approach that checks every plan in the
reachable belief space $\reachable_h(b_{\textrm{init}})$ requires
$O(|\operator|^h|\observation|^h)$ calls to the SMT solver. This
exponential growth of the reachable belief space
$\reachable_h(b_{\textrm{init}})$ due to branches on both action and
observations is a major challenge for synthesizing a \emph{valid}
policy.

In our case, \ipsAlgo{} exploits the notion of \emph{\satBS{}}
$\reachable^*(b_{\textrm{init}}, \objective)$ and efficiently explores the
\emph{\satBS{}} $\reachable^*(b_{\textrm{init}}, \objective)$ by leveraging an
incremental SMT solver to generate a candidate plan $\sigma$ of length
at most $h$. This candidate plan fixes the choice of actions at each
step and thus the policy generation process only needs to consider the
branches on observations for each step, as shown in
\figref{} \ref{fig:ips_ex}. Therefore, \ipsAlgo{} requires
$O(I|\observation|^h)$ calls to the SMT solver, where $I$ is the
number of interactions between plan generation and policy generation,
while the naive approach described above requires
$O(|\operator|^h|\observation|^h)$ SMT solver calls. In general, $I$
is often much smaller than $|\operator|^h$, which leads to much faster
policy synthesis. Therefore, we expect our method to be effective for
POMDPs with a high-dimensional action space and a restricted partially
observable component, but would not scale well for POMDPs with
high-dimensional/continuous observation space.

\section{Experiments}
We evaluate \ipsAlgo{} in a partially observable kitchen domain
(\figref{} \ref{fig:kitchen}) with a PR2 robot and $M$ uncertain
obstacles placed in the yellow ``shadow'' region. The task for the
robot is to safely pass the yellow ``shadow'' region avoiding
collisions with uncertain obstacles and eventually pick up a green cup
from the black storage area.

We first discretize the kitchen environment into $N$ \emph{cell}s. We
assume that the locations of the obstacles are uniformly distributed
among the cells in the yellow ``shadow'' region and there is at most
one obstacle in each cell. We also assume the robot starts at a known
initial location. However, due to the robot's imperfect perception,
the locations of the robot, the locations of uncertain obstacles, and
the location of the target cups are all partially observable during
execution.

In this domain, the actuation and perception of the robot are
imperfect. There are ten uncertain robot actions ($|\operator| = 10$):
\begin{enumerate}
\item Four \emph{move} actions that move the robot to an adjacent cell
  in four directions: including \emph{move-north}, \emph{move-south},
  \emph{move-west} and \emph{move-east}. \emph{Move} actions could
  fail with a probability $p_{\textrm{fail}}$, resulting in no change in the
  state.
\item Four \emph{look} actions that observe a cell to see whether
  there is an obstacle in that cell, including \emph{look-north},
  \emph{look-south}, \emph{look-west}, \emph{look-east} (look at the
  adjacent cell in the corresponding direction). When the robot calls
  \emph{look} to observe a particular $\textrm{cell}_i$, it may either
  make an observation $o = o_{\textrm{pos}}$ representing the robot
  observes an obstacle in $\textrm{cell}_i$ or $o = o_{\textrm{neg}}$
  representing the robot observers no obstacle in
  $\textrm{cell}_i$. The probabilistic observation function
  $\observemodel(s', a, o)$ for \emph{look} actions is defined based
  on the false positive probability $p_{\textrm{fp}}$ and the false
  negative probability $p_{\textrm{fn}}$.
\item Two \emph{pick-up} actions that pick up an object from the black
  storage area: pick-up using the left hand $a_L$ and pick-up using
  the right hand $a_R$. The model of \emph{pick-up} actions is the
  same as what we discussed in \secref{} \ref{sec_constrainedPOMDPs}
  (see \figref{} \ref{fig_pomdp}).
\end{enumerate}

The task shown in \figref{} \ref{fig:kitchen} can be specified as a
safe-reachability objective with a set $\dest$ of goal beliefs
and a set $\safe$ of safe beliefs, defined as follows:
\begin{align}
  \dest & = \{b \in \beliefspace~|~\left(\sum b(\textrm{target cup in
          robot's hand}))\right) > 1 - \delta_1\}
          \nonumber\\
  \safe & = \{b \in \beliefspace~|~\left(\sum b(\textrm{robot in collision}))\right) < \delta_2\}
\end{align}
where $\delta_1$ and $\delta_2$ are small values that represent
tolerance. The reachability objective specifies that in a goal belief,
the probability of having the target cup in the robot's hand should be
greater than the threshold $1 - \delta_1$. The safety objective
specifies that in a safe belief, the probability of the robot in
collision (the robot and one obstacle in the same cell) should be less
than the tolerance $\delta_2$.



We evaluate the performance of \ipsAlgo{} using test cases of the
kitchen domain with various numbers of obstacles. We use Z3
\cite{de2008z3} as our backend incremental SMT solver. All experiments
were conducted on a 3.0GHz Intel{\small
  \textsuperscript{\textregistered}} processor with 32GB memory. For
all the tests, the horizon bound is $h = 20$ and the number of cells
in the kitchen environment is $N = 24$.

\def\plotwidth{.99\columnwidth}
\begin{figure}[t!]
    \centering
    \begin{tikzpicture}[tight background]
      \begin{axis}[height=2.5 in, width=\plotwidth, legend
        style={legend pos=north west, xshift=0.25em}, ylabel
        style={xshift=-0.75em}, ylabel= Policy Synthesis Time (s),
        ymode=log,xlabel=Number of obstacles $M$, xtick={1,...,4},
        xmin=0.75, xmax=4.25]
        
        \addplot[mark=o,color=black,only marks,thick]
        plot file {data/computation_time_bps_with_inc.dat};
        
        \addplot[mark=square,color=black,only marks,thick]
        plot file {data/computation_time_bps_no_inc.dat};
        
        
        \addlegendentry{BPS (with inc.)}
        \addlegendentry{BPS (no inc.)}
    \end{axis}
    \end{tikzpicture}
    \caption{Performance of \ipsAlgo{} 
      as the number of
      obstacles $M$ varies. The plot of circles shows the performance of
      BPS with incremental solving and the plot of squares shows the
      performance of BPS without incremental solving.
    }
    \label{fig:time}
\end{figure}
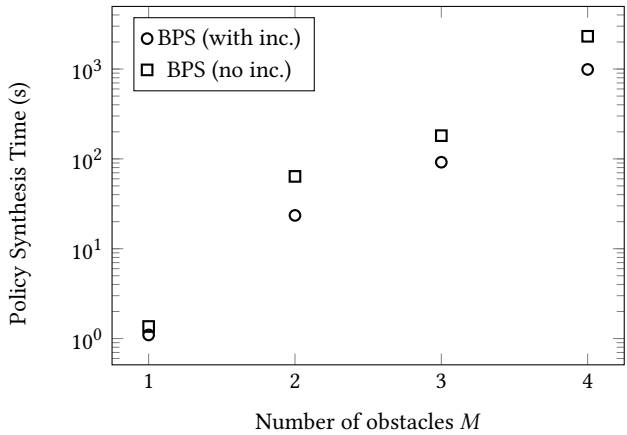


To evaluate the gains from incremental solving, we test \ipsAlgo{} in
two settings: with and without incremental solving. Note that when
incremental solving is disabled, each call to the SMT solver requires
solving the SMT constraints from scratch, rather than reusing the
results from the previous SMT solver calls. \figref{} \ref{fig:time}
shows the performance results of \ipsAlgo{} with and without
incremental solving. As we can see from \figref{} \ref{fig:time},
enabling incremental solving in BPS leads to a performance improvement
in policy synthesis. This is because the BMC encoding
\cite{biere2003bounded} used in \ipsAlgo{} is particularly suitable
for incremental solving since increasing horizon and blocking invalid
plans correspond to pushing/popping constraints.

\begin{figure}[t!]
    \centering
    \begin{tikzpicture}[only marks, tight background]
      \begin{axis}[height=2.5 in, width=\plotwidth, ylabel
        style={xshift=-0.75em}, ylabel=Number of Plans Checked,
        ytick={0,20,...,140},
        xlabel=Number of obstacles $M$, xtick={0,1,...,5}]
        \addplot[mark=o,color=black,thick]
        plot file {data/num_plans_checked_bps.dat};
    \end{axis}
    \end{tikzpicture}
    \caption{The number of plans checked (i.e, the number of SMT
      calls) by BPS during policy synthesis as the number of obstacles
      $M$ varies.}
    \label{fig:plan}
\end{figure}
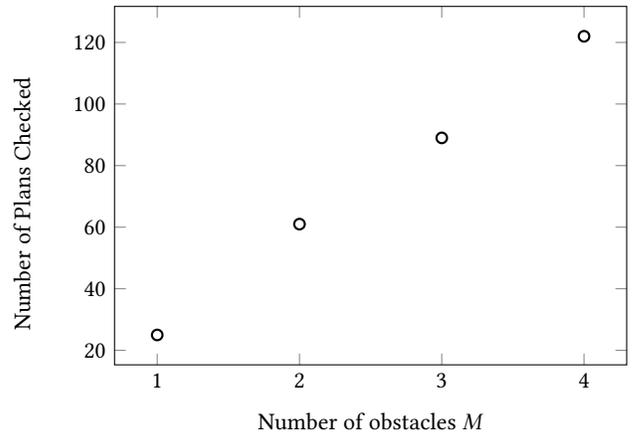
   
To demonstrate the gains from utilizing the \satBS{} compared to a
naive exhaustive search in the reachable belief space, we first
estimate the number of plans in the reachable belief space. There is
no observation branching for the four \emph{move} actions and there
are two observation branches for the four \emph{look} actions. We
ignore the two \emph{pick-up} actions since these two actions are not
available in every step and can only be performed when the robot is
fairly confident that it is in the position where it is ready to pick
up a cup from the black storage area. Therefore, the approximate lower
bound of the number of plans in the reachable belief space with at
most $h = 20$ steps is $(4 + 4\times 2)^{20} \approx
10^{21}$. However, as we can see from \figref{} \ref{fig:plan} where
we show the number of plans checked by BPS during policy synthesis,
for the largest test, the number of plans checked (around 120) in
\ipsAlgo{} is very small compared to the number of plans in the
reachable belief space. These results show that \ipsAlgo{} can solve
problems in huge reachable belief spaces with a small number of SMT
solver calls by focusing on the \satBS{}.

However, \figref{} \ref{fig:time} also shows that the synthesis time
grows exponentially as the number of obstacles increases, which
matches our complexity analysis in \secref{}
\ref{sec:complexity}. This is because the current implementation of
\ipsAlgo{} operates on an \emph{exact} tree representation of policies
with all the observation branches. As the number of obstacles
increases, both the horizon bound (the height of the policy tree) and
the size of the state space (the belief space dimension) increase,
which leads to an exponential growth of plans in the policy tree and
makes the policy synthesis problem much harder.

\section{Conclusion and Discussion}

We present a novel policy synthesis method called \ipsAlgo{} for
POMDPs with safe-reachability objectives. We exploit the notion of a
\satBS{} to improve computational efficiency. We construct constraints
in a way similar to Bounded Model Checking \cite{biere2003bounded} to
compactly represent the \satBS{}, which we efficiently explore through
an incremental Satisfiability Modulo Theories solver
\cite{de2008z3}. We evaluate \ipsAlgo{} in an uncertain robotic domain
and the results show that our method can synthesize policies for large
problems by focusing on the \satBS{}.

The current implementation of \ipsAlgo{} operates on an \emph{exact}
representation of the policy (the tree structure shown in \figref{}
\ref{fig:ips_ex}). As a result, \ipsAlgo{} suffers from the
exponential growth as the horizon increases. An important ongoing
question is how to approximately represent the policy while preserving
correctness. Another issue arises from the discrete representations
(discrete POMDPs) used in our approach. While many robot tasks can be
modeled using these representations, discretization often suffers from
the ``curse of dimensionality''. Investigating how to deal with
continuous state spaces and continuous observations directly without
discretization is another promising future direction for this work and
its application in robotics.

\section{Acknowledgments}
This work was supported in part by NSF CCF 1139011, NSF CCF 1514372,
NSF CCF 1162076 and NSF IIS 1317849.


\bibliographystyle{ACM-Reference-Format}  
\bibliography{references}  

\end{document}